\documentclass[table, 11pt]{article}
\usepackage[table, svgnames, dvipsnames]{xcolor}

\usepackage[final]{acl}

\usepackage{times}
\usepackage{latexsym}

\usepackage[T1]{fontenc}

\usepackage[utf8]{inputenc}
\usepackage{comment}

\usepackage{microtype}

\usepackage{inconsolata}
\usepackage{hyperref}

\usepackage{graphicx}
\usepackage{amsmath}

\usepackage{colortbl}

\definecolor{lightgray}{gray}{0.80}

\newcolumntype{g}{>{\columncolor{lightgray}}c}

\title{Preserving Generalization of Language Models in Few-shot Continual Relation Extraction}

\author{
  \textbf{Quyen Tran\textsuperscript{1}\footnotemark[1]},
  \textbf{Thanh Nguyen\textsuperscript{2}\footnotemark[1]},
  \textbf{Anh Nguyen\textsuperscript{2}\footnotemark[1]},
  \textbf{Nam Le\textsuperscript{2}}, \\
  \textbf{Trung Le\textsuperscript{3}},
  \textbf{Linh Ngo Van \textsuperscript{2}\footnotemark[2]},
  \textbf{Thien Huu Nguyen\textsuperscript{4}}
  \bigskip \\
\textsuperscript{1}VinAI Research,
\textsuperscript{2}Hanoi University of Science and Technology, \\
\textsuperscript{3}Monash University, 
\textsuperscript{4}University of Oregon
}

\begin{document}
\maketitle

\renewcommand{\thefootnote}{\fnsymbol{footnote}}
\footnotetext[1]{Equally contributed.}
\footnotetext[2]{Corresponding author: \href{mailto:email@domain}{linhnv@soict.hust.edu.vn}}
\renewcommand*{\thefootnote}{\arabic{footnote}}

\begin{abstract}

Few-shot Continual Relations Extraction (FCRE) is an emerging and dynamic area of study where models can sequentially integrate knowledge from new relations with limited labeled data while circumventing catastrophic forgetting and preserving prior knowledge from pre-trained backbones. In this work, we introduce a novel method that leverages often-discarded language model heads. By employing these components via a mutual information maximization strategy, our approach helps maintain prior knowledge from the pre-trained backbone and strategically aligns the primary classification head, thereby enhancing model performance. Furthermore, we explore the potential of Large Language Models (LLMs), renowned for their wealth of knowledge, in addressing FCRE challenges. Our comprehensive experimental results underscore the efficacy of the proposed method and offer valuable insights for future work.

\end{abstract}

\section{Introduction}

Continual Relations Extraction (CRE) is a learning scenario that requires a model to identify emerging relationships between entities or objects in texts  while maintaining the accuracy of existing classifications and avoiding the problem of \textit{Catastrophic forgetting} \citep{THRUN199525, DBLP:journals/neco/FrenchC02}. In many real-world situations, models must learn from a few new samples due to the limited availability of labeled training data for relations. As a result, Few-short Continual Relation Extraction (FCRE) methods have been proposed \citep{DBLP:conf/acl/QinJ22, DBLP:conf/acl/ChenWS23} to enable models to solve new tasks where each new relation has only a minimal number of corresponding samples. However, due to the lack of data, FCRE models are often biased towards the current task compared to related scenarios, which can lead to forgetting previous knowledge and losing highly general priori from the pre-trained backbone. Thus, the challenge of FCRE is not only catastrophic forgetting but also severe overfitting.

\begin{figure}[t]
    \centering
    \includegraphics[width=\columnwidth]{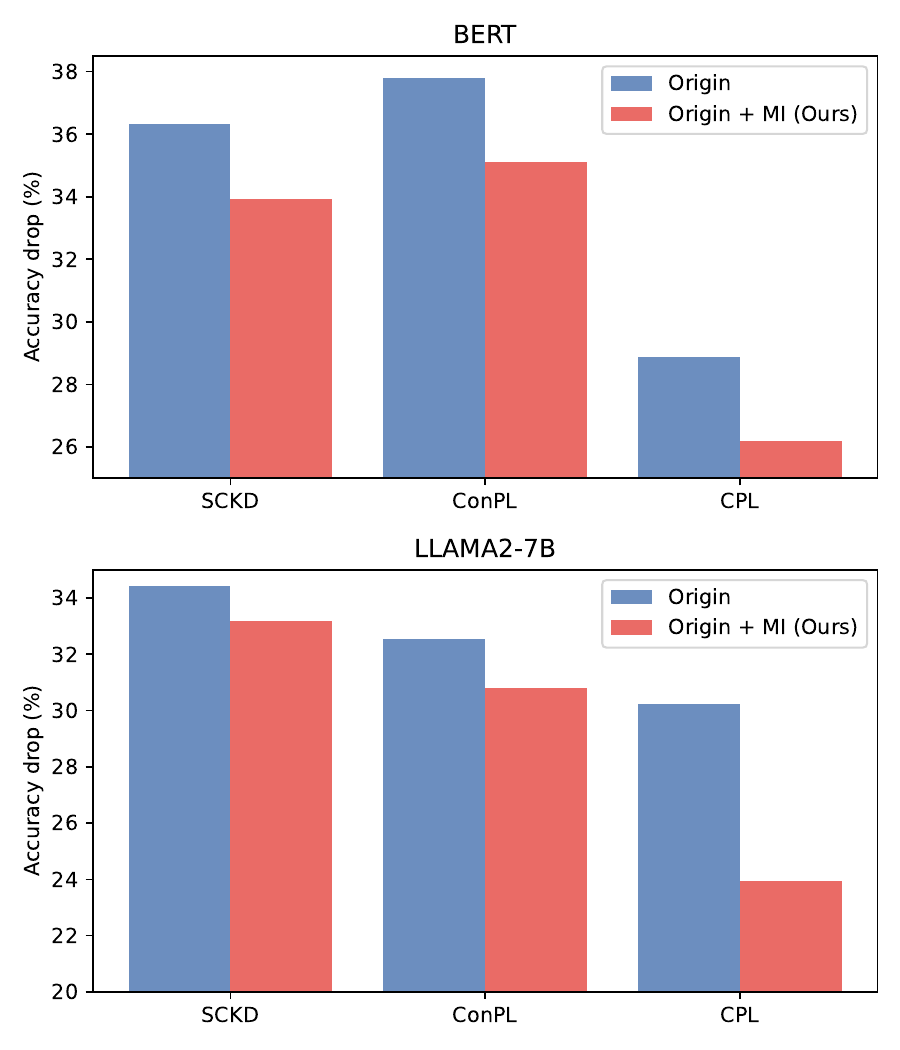}
    \vspace{-5mm}
    \caption{Accuracy drop (\%) after learning eight tasks of methods on TACRED \textit{5-way-5-shot}. Lower is better.}
    \label{fig:forget}
    \vspace{-4mm}
\end{figure}

\begin{figure}[ht]
    \centering
    \includegraphics[width=1.\columnwidth]{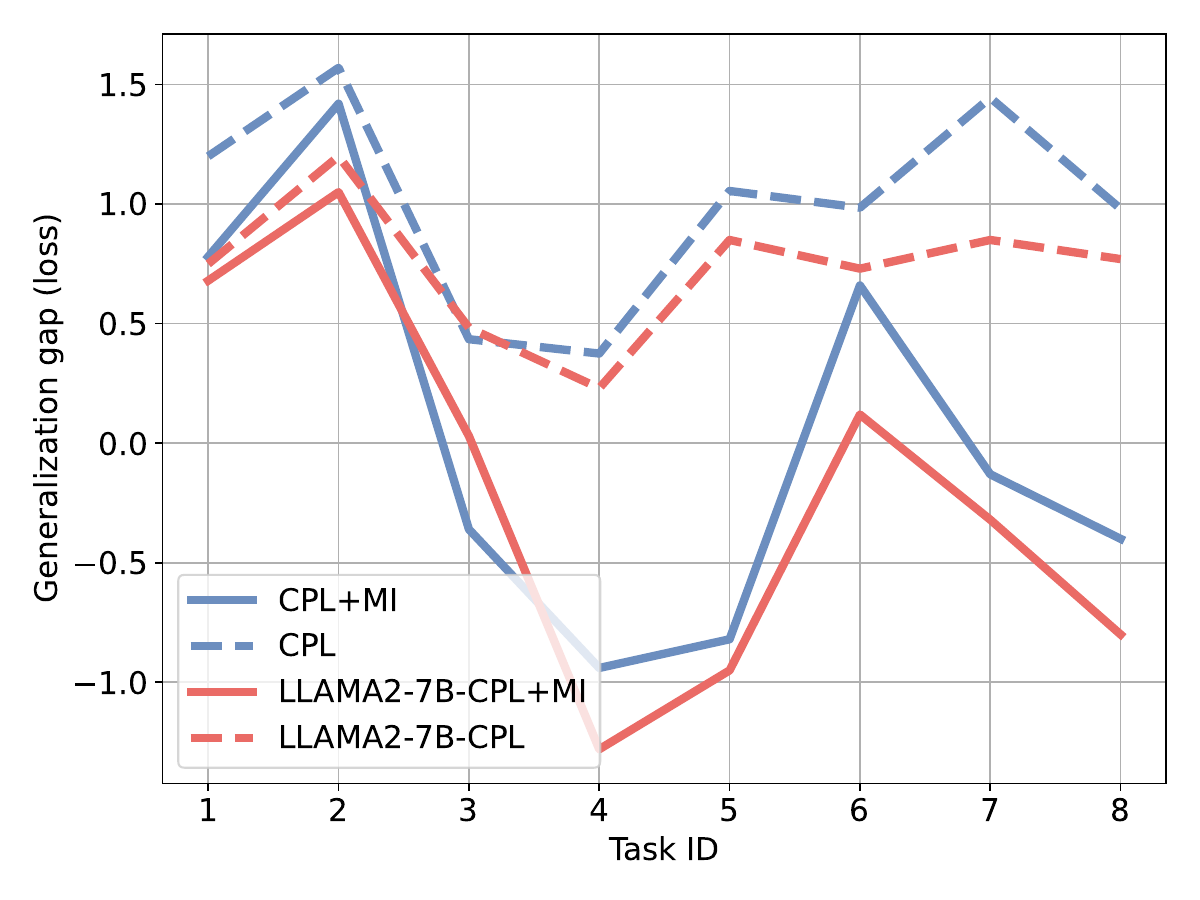}
    \vspace{-6mm}

    \caption{Generalization gap regarding loss of models after training each task (TACRED \textit{5-way-5-shot}, seed=100).}
    \label{fig:overfit}
    \vspace{-5mm}
\end{figure}


Recent works \citep{DBLP:conf/acl/WangWH23, DBLP:conf/acl/QinJ22, DBLP:conf/acl/ChenWS23} tackles these issues by employing memory-based approaches inspired by traditional Continual Learning methods \cite{DBLP:conf/nips/RolnickASLW19, DBLP:conf/nips/BuzzegaBPAC20, DBLP:conf/nips/Lopez-PazR17}, along with various strategies to enhance the model's ability to distinguish relation representations. Nevertheless, these methods solely fine-tune pre-trained BERT-based backbones for few-shot tasks, which leads to eroding prior knowledge from the pre-trained model and hindering the final performance. {Additionally, these methods often neglect the pre-trained LM head in favor of training a new classifier from scratch, even though this component contains rich and general knowledge that remains untapped. Therefore, we propose our \textit{Mutual Information Maximization (MIM) strategy} that leverages pre-trained LM heads during training FCRE models for the first time. Our proposed strategy not only helps preserve the knowledge on the backbone but also assists in aligning the main classifier to improve representation learning. Extensive experimental results on benchmark datasets demonstrate the effectiveness of our novel approach in preserving the pre-trained LM's generalization capability and reducing forgetting, leading to remarkable results.

Furthermore, pre-trained Large Language Models (LLMs) \citep{DBLP:journals/corr/abs-2307-09288, jiang2023mistral} with billions of parameters are known for their excellence in autoregressive text generation tasks. They have also been extensively studied in text classification and information extraction \citep{DBLP:conf/icml/ZhaoWFK021, DBLP:journals/corr/abs-2302-10205}. However, these models often underperform compared to discriminative encoder models like BERT due to their generation-focused mechanism. To address this, recent work \citep{DBLP:journals/corr/abs-2310-01208} proposed replacing ineffective LLM heads with classification heads in the restricted space of the classification problem. This approach has shown promise, but the potential of LLMs in CL, specifically in FCRE, remains underexplored. Therefore, we conduct extensive experiments to answer: How the performance would LLMs yeild for FCRE? How will limited data in this scenario impact the generalization of LLMs? We also assess the effectiveness of our MIM strategy when using LLM heads, which were eliminated due to their unsuitability. The results offer valuable insights for the community.

To sum up, our main contributions are twofold:
\begin{itemize}
    \item First, we introduce a novel approach to enhance FCRE models by strategically leveraging the LM heads. {{Through maximizing}} mutual information between these components and the primary classifiers, we can better preserve prior knowledge from pre-trained backbones, as well as strengthen representation learning. The experimental results demonstrate our effectiveness.

    \item We also investigate the application of pre-trained LLMs to FCRE tasks, including evaluating the effectiveness of the proposed method when using LLM heads, which were discarded in classification-based problems due to their unsuitability. Our comprehensive experimental results offer valuable insights.

\end{itemize}

\section{Related work}
\paragraph{Continual Learning (CL)}is a learning scenario that requires models to continually acquire new knowledge from a sequence of tasks while preventing the loss of previously learned information. The main challenge in CL is \textit{catastrophic forgetting} \citep{DBLP:conf/nips/French93}. To address this problem, memory-based approaches prove to be effective methods for both machine learning \citep{DBLP:conf/cvpr/RebuffiKSL17, DBLP:conf/nips/ShinLKK17} and NLP problems \citep{DBLP:conf/naacl/WangXYGCW19, DBLP:conf/acl/HanDGLLLSZ20}. In particular, models need to save a few representative samples from the current task in a memory buffer and replay these samples when learning new tasks to review old knowledge. 

\paragraph{Fewshot Continual Relation Extraction} is a challenging scenario, which was introduced by \citep{DBLP:conf/acl/QinJ22} for Relation Extraction problems.  This challenge arises due to the limited availability of data for new tasks, coupled with the high cost and time involved in obtaining high-quality data. Recent work like \citet{DBLP:conf/acl/WangWH23, DBLP:conf/acl/ChenWS23, DBLP:conf/coling/MaHL024} propose memory-based solutions, which suggest imposing objective functions on the embedding space and classification head. Specifically, \citet{DBLP:conf/acl/WangWH23} employs serial objective functions based on contrastive and distillation,  \citet{DBLP:conf/acl/QinJ22} leverage extra training data from unlabeled text, and \citet{DBLP:conf/acl/ChenWS23} proposes a consistent prototype learning strategy to help the model distinguish between different relation representations, thus enhancing representation learning efficiency. 

However, in these methods, eliminating the pre-trained LM head and training a new classifier still leads to overfitting and forgetting due to limited data, as it emphasizes discriminative features only. To address this problem, we propose a novel approach that leverages LM heads, which are often overlooked in pre-trained models for downstream tasks. Our method not only helps preserve prior knowledge from the backbone but also supports the training of the main classifier, thereby further reducing both catastrophic forgetting and overfitting.

\section{Background}
\subsection{Problem Formulation}

In the setting of FCRE, a model needs to continually acquire new knowledge from a series of tasks. For each task $t$, also denoted as $\mathcal{T}^t$, the model is trained on the training set $D^t = \{ (x^t_i, y^t_i)\}_{i=1}^{N\times K}$. Here, $N$ and $K$ represent the number of classes in the new relation set $R^t$ and the number of samples corresponding to each relation, respectively. Each sample $(x^t_i, y^t_i)$ consists of a sentence $x_i$ with a pair of entities $(e_h, e_t)$ and a relation label $y_i \in R^t$. This type of task is also known as \textit{"N-way-K-shot"}. Once task $\mathcal{T}^t$ is completed, $D^t$ is no longer available for future learning. Finally, the model will be evaluated on all task data so far in order to identify relations in $\tilde{R^t} = \bigcup_{i=1}^t R^i$. 


\subsection{Existing Concept of FCRE Models}



Current FCRE methods \citep{DBLP:conf/acl/WangWH23, DBLP:conf/acl/ChenWS23, DBLP:conf/coling/MaHL024} have considered tackling two main issues: catastrophic forgetting and overfitting. This has been achieved by exploiting the power of pre-trained BERTs and various motivated techniques which can divided into 3 main groups, including (i) using objective functions (i.e., $\mathcal{L}_0$) to enhance representation learning ability, (ii) implementing a prompt design, and (iii) employing a memory management strategy to store and retrieve knowledge of old tasks. In this paper, we propose a novel strategy that can flexibly integrate with and improve these methods (Figure \ref{fig:framework}). 




Moreover, to explore the potential of pre-trained LLMs when dealing with the FCRE problems, we need to apply the current SOTA methods for LLMs, which were originally designed for "encoder-only" models. On the other hand, the examined LLMs (LLAMA2, Mistral) are "decoder-only", operating in the auto-regressive mechanism \citep{DBLP:journals/corr/abs-1711-09534, DBLP:journals/corr/abs-1906-08237}. Due to the differences between these models, we have to modify the original designs mentioned above (see Sec. \ref{LLM_invest}).



\section{Proposed Method}
{In this section, we first present our efficient strategy in Section \ref{MIM_strategy} that can flexibly adapt to the existing FCRE methods and enhance model performance. After that, in Section \ref{LLM_invest}, we explain in detail the motivation and research questions when investigating LLMs in FCRE.}






\begin{figure}[ht]
\includegraphics[width=1.\columnwidth]{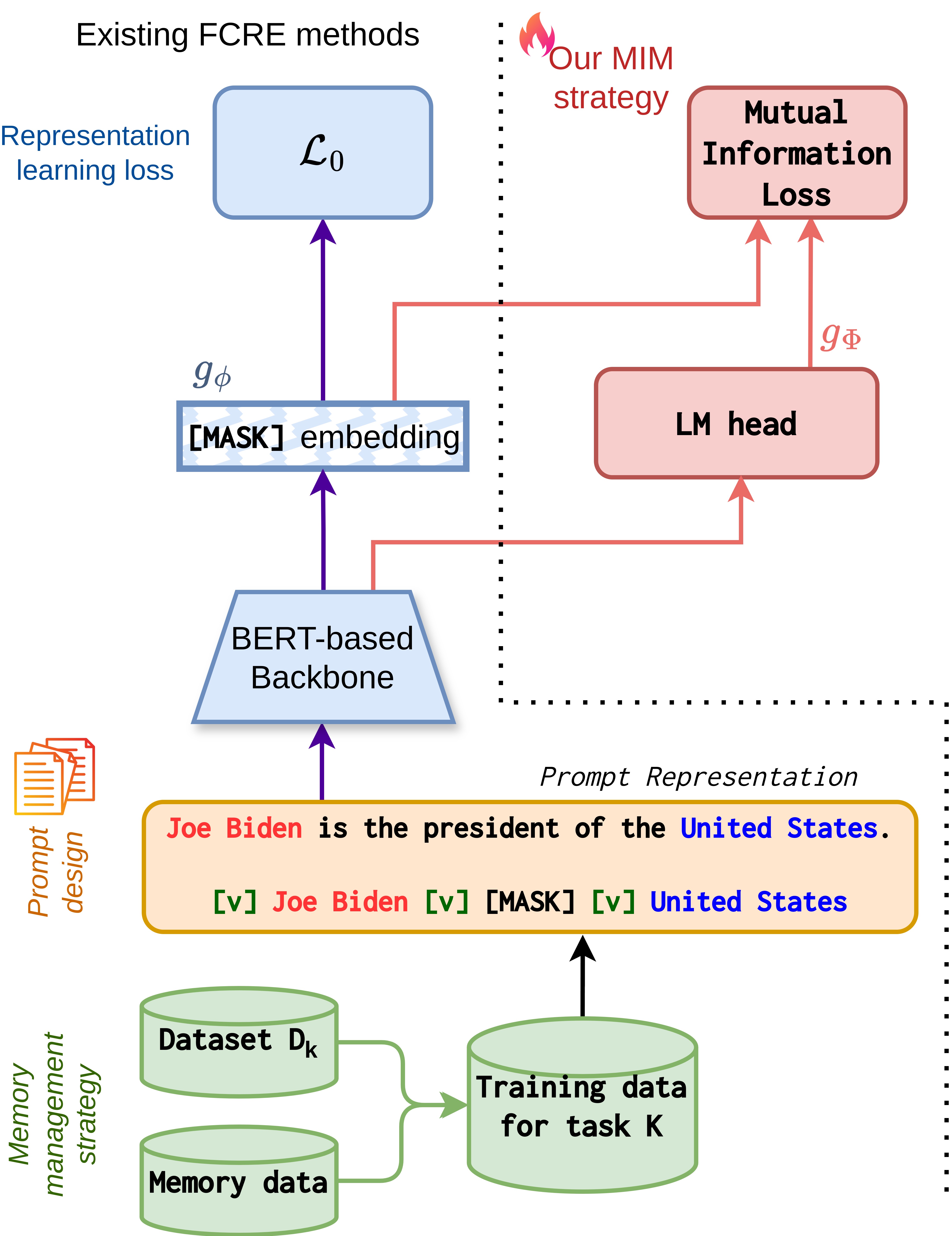}
  \caption{Our Framework}
  \label{fig:framework}
    \vspace{-4mm}
\end{figure}


\subsection{Mutual Information Maximization (MIM)}
\label{MIM_strategy}


\begin{table*}[!ht]
    \centering

    \resizebox{.8\textwidth}{!}{
    \begin{tabular}{lcccccccl}
    \multicolumn{9}{l}{\textbf{FewRel} \textit{(10-way 5-shot)}} \\
    \hline
        Method & $\mathcal{T}^1$ & $\mathcal{T}^2$ & $\mathcal{T}^3$ & $\mathcal{T}^4$ & $\mathcal{T}^5$ & $\mathcal{T}^6$ & $\mathcal{T}^7$ & $\mathcal{T}^8$ \\ \hline \hline
        SCKD & 94.75 & 82.83 & 76.21 & 72.19 & 70.61 & 67.15 & 64.86 & 62.98  \\ 
        \rowcolor{lightgray} SCKD+MI & \textbf{94.75} & \textbf{83.88} & \textbf{76.71} & \textbf{72.34} & \textbf{70.78} & \textbf{67.36} & \textbf{65.08} & \textbf{63.95} \small \textcolor{red}{ $\uparrow$ 0.97} \\ \hline \hline
        ConPL$^{**}$  & \textbf{95.18} & 79.63 & 74.54 & 71.27 & 68.35 & 63.86 & 64.74 & 62.46 \\ 
        \rowcolor{lightgray} ConPL+MI & 95.02 & \textbf{81.42} & \textbf{77.23} & \textbf{74.21} & \textbf{69.64} & \textbf{67.74} & \textbf{66.44} & \textbf{64.50} \small \textcolor{red}{ $\uparrow$ 2.04}\\ \hline \hline
        CPL & \textbf{94.87} & 85.14 & 78.80 & 75.10 & 72.57 & 69.57 & 66.85 & 64.50 \\ 
        \rowcolor{lightgray} CPL+MI & 94.69 & \textbf{85.58} & \textbf{80.12} &\textbf{ 75.71} & \textbf{73.90} & \textbf{70.72} & \textbf{68.42} & \textbf{66.27} \small \textcolor{red}{ $\uparrow$ 1.77} \\ \hline \\ 
        \multicolumn{9}{l}{\textbf{TACRED} \textit{(5-way 5-shot)}} \\ \hline
        Method & $\mathcal{T}^1$ & $\mathcal{T}^2$ & $\mathcal{T}^3$ & $\mathcal{T}^4$ & $\mathcal{T}^5$ & $\mathcal{T}^6$ & $\mathcal{T}^7$ & $\mathcal{T}^8$ \\
        \hline
        \hline
        SCKD & \textbf{88.42} & 79.35 & 70.61 & \textbf{66.78} & 60.47 & 58.05 & 54.41 & 52.11 \\ 
        \rowcolor{lightgray} SCKD+MI & 87.55 & \textbf{79.39} & \textbf{70.70 }& {66.68} & \textbf{61.94} & \textbf{59.81} & \textbf{55.10} & \textbf{53.63} \small \textcolor{red}{ $\uparrow$ 1.52}\\ 
        \hline \hline
        ConPL$^{**}$ & \textbf{88.77} & 69.64 & 57.50 & 52.15 & 58.19 & 55.01 & 52.88 & 50.97 \\ 
        \rowcolor{lightgray} ConPL+MI & 88.10 & \textbf{83.03} & \textbf{73.19} & \textbf{65.21} & \textbf{59.77} & \textbf{60.99} & \textbf{58.88} & \textbf{52.98} \small \textcolor{red}{ $\uparrow$ 2.01} \\
        \hline \hline
        CPL & \textbf{86.27} & 81.55 & 73.52 & 68.96 & 63.96 & 62.66 & 59.96 & 57.39 \\ 
        \rowcolor{lightgray}CPL+MI  & 85.67 & \textbf{82.54} & \textbf{75.12} & \textbf{70.65} & \textbf{66.79} & \textbf{65.17} &\textbf{ 61.25} & \textbf{59.48} \small \textcolor{red}{ $\uparrow$ 2.09} \\ \hline
        
    \end{tabular}}
    \caption{Accuracy (\%) of different BERT-based methods after training for each task on TACRED and FewRel in \textit{5-shot} settings. We \colorbox{lightgray}{highlight} the rows corresponding to our method. The best result in each group is in \textbf{bold}. **Results of ConPL are reproduced (see Section \ref{exp_setup})}
    
    \label{table:main1}
    \vspace{-4mm}
\end{table*}


According to recent work \cite{DBLP:journals/corr/abs-2310-01208, xu2023large}, using pre-trained LMs (BERTs) with their classification heads often leads to poor results. This is because the models must return responses in the vocabulary's high-dimensional space (i.e., $\| V \|$). Therefore, in downstream tasks like Relation Extraction, LM heads of pre-trained LMs are often discarded. Instead, existing work \citep{DBLP:conf/acl/WangWH23, DBLP:conf/coling/MaHL024} opt for training a classification head across tasks as a better solution. However, in FCRE, training a new classifier from scratch often encourages models to emphasize only discriminative features derived from sparse data streams and memory buffers. This biased behavior can make the model seriously overfit and rapidly lose prior knowledge from the pre-trained backbone and, thus, hinder the final performance.



Therefore, we propose an MIM strategy that exploits the overlooked LM head to solve the drawbacks of existing FCRE methods. Intuitively, leveraging knowledge from pre-trained LM heads will support the primary classifier, aiding the model in capturing information more holistically and better preserving old knowledge of the pre-trained backbone. 
In particular, inspired by \cite{pmlr-v162-guo22g}, we aim at maximizing Mutual Information \textit{(MI)} between latent representations on the LM head branch and on our main classifier branch as follows:

\begin{equation}
 MI = I[g_\phi(\mathbf{x}), g^{LM}_\Phi(\mathbf{x})]   
\end{equation}
where $g_\phi$ corresponds to the class-discriminative feature representation at the classification head,  $g^{LM}_\Phi$ denotes the representation at the LM head. According to \cite{DBLP:journals/corr/abs-1807-03748}:
\begin{equation}
    MI \geq \log B + \textnormal{InfoNCE} (\{ x_i\}_{i=1}^B; h)
\end{equation}
where we have defined
\begin{equation}
    \begin{split}
        & \textnormal{InfoNCE} (\{ x_i\}_{i=1}^B; h) =  \\
        & \frac{1}{B}\sum_{i=1}^B \log \frac{h(g_\phi(\mathbf{x_i}), g^{LM}_\Phi(\mathbf{x_i}))}{\sum_{j=1}^B h(g_\phi(\mathbf{x_i}), g^{LM}_\Phi(\mathbf{x_j}))}, \\
        & h(g_\phi(\mathbf{x_i}), g^{LM}_\Phi(\mathbf{x_j})) = \exp \frac{g_\phi(\mathbf{x_i})^TWg^{LM}_\Phi(\mathbf{x_j})}{\tau}
    \end{split}
    \end{equation}
    where $\tau$ is the temperature, $B$ is mini-batch size and $W$ is a trainable parameter. Then, the MI loss function in our implementation is: 
\begin{equation}
    \mathcal{L}_{MI} = -\sum_{(x_i , y_i) \in D_{train}^{k}}\textnormal{InfoNCE} (\{x_i\}_{i=1}^B; h)
\end{equation}

Therefore, the objective function of the model can be summarized as:
\begin{equation}
  \mathcal{L} = \mathcal{L}_{0} + \mathcal{L}_{MI} 
\end{equation}
where $\mathcal{L}_{0}$ is the loss function of the original method. In this work, to demonstrate the effectiveness of our method, we integrate it into three existing methods CPL \cite{DBLP:conf/coling/MaHL024}, ConPL \cite{DBLP:conf/acl/ChenWS23} and SCKD \cite{DBLP:conf/acl/WangWH23} (see Appendix \ref{sec:appendix_baseline}).

\paragraph{Discussion:} Although using pre-trained LM heads directly in downstream tasks is challenging, this does not hinder us from tapping into their wealth of knowledge to enhance our model performance in FCRE. 
\begin{itemize}
    \item First, maintaining the LM heads while fine-tuning them with a carefully controlled learning rate encourages the pre-trained backbones to retain prior knowledge and inherent behaviors. Thus, this strategy can mitigate the risk of overfitting, especially when models are trained on limited data for each task, enhancing their overall robustness and reliability.

    \begin{figure}[ht]
  \includegraphics[width=1.\columnwidth]{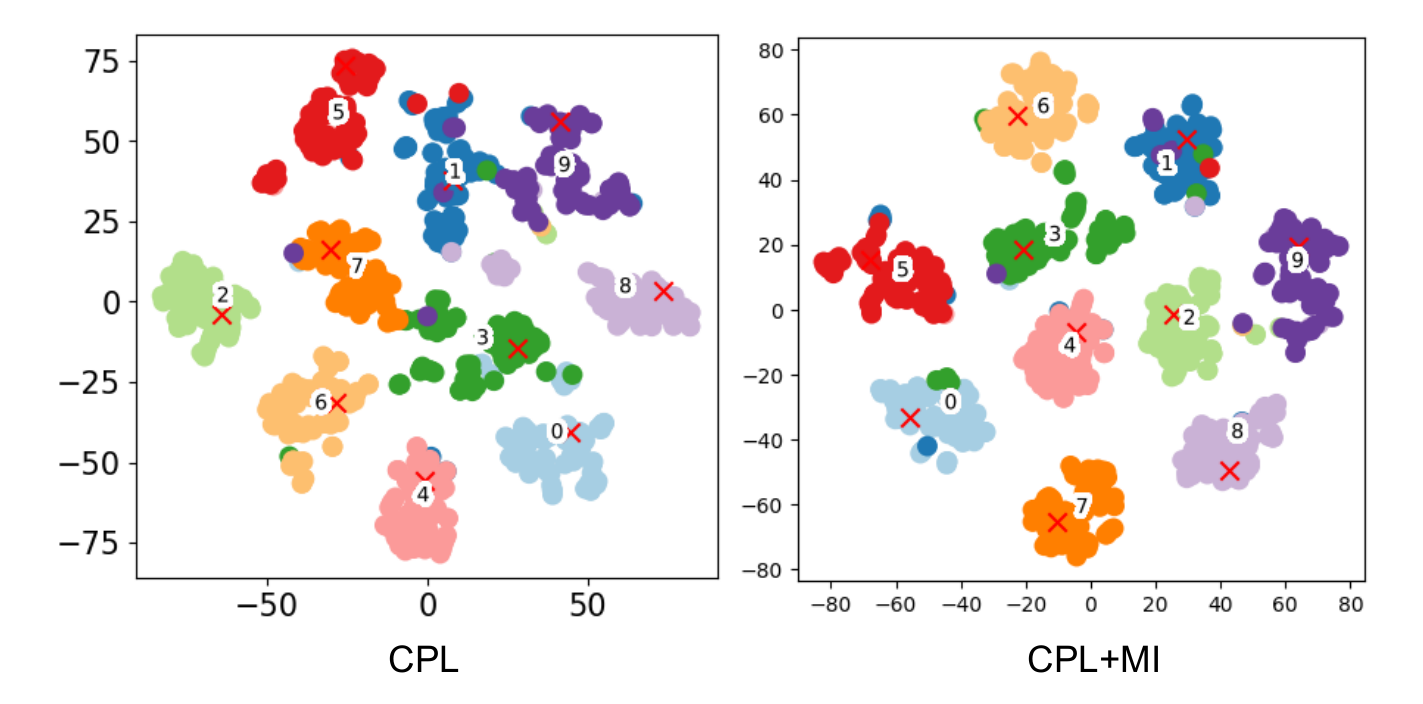}
  \vspace{-6mm}
  \caption{\textit{t}-SNE visualization for representation of 10 relations from Task 1 on the main classification branch after the last task (FewRel \textit{10 way - 5shot}).}
  \label{fig:overlap_main}
    \vspace{-3mm}
\end{figure}

    \item Second, applying MIM on different representation layers of the data will be a powerful aid for $\mathcal{L}_0$ in learning representations. Specifically, the mutual information of samples with the same label will be enhanced, while the information corresponding to features of different labels will be restricted. As a result, feature vectors of the same class will become more condensed, and representations of different classes will be more separated.
\end{itemize}







\subsection{Exploiting LLMs for FCRE}
\label{LLM_invest}

\paragraph{Motivations and Research questions} Pre-trained LLMs \citep{DBLP:journals/corr/abs-2307-09288, jiang2023mistral} are known for containing rich knowledge with billions of parameters, which have achieved impressive results in auto-regressive text generation tasks. These models have also been extensively examined in classification-based problems \citep{DBLP:conf/icml/ZhaoWFK021, DBLP:journals/corr/abs-2302-10205}. However, these models often do not outperform discriminative encoder models such as BERT because their original generation-focused mechanism, which generates answers over a large vocabulary, may not capture task-specific patterns as efficiently as label-supervised BERT models. To address this drawback, recent work \cite{DBLP:journals/corr/abs-2310-01208} proposed directly extracting latent representations from the final LLaMA decoder layer and mapping them into the label space through feed-forward layers. Specifically, the LLM heads, which have been found ineffective, are removed and replaced by a classification head trained from scratch using CrossEntropy loss. This approach has shown promising results. However, exploration in the area of Continual Learning, specifically Few-shot Continual Relation Extraction (FCRE), has not yet been thoroughly investigated. Therefore, in this work, we conduct extensive experiments to answer the following research questions  \textit{\textbf{(RQs)}}: 



\begin{itemize}
    \item \textit{\textbf{RQ1:} How the performance would LLMs yield in FCRE tasks?} Will it yield significantly better results compared to conventional BERT-based models? How will the limited data in the FCRE scenario impact the generalization of this model class? It would be interesting to examine the behavior of an LLM, which contains rich prior knowledge in the context of the FCRE problem, where each task only has very little data, and the model will usually be forgotten and severely overfit.


    \item \textit{\textbf{RQ2:}} Our study also aims to assess \textit{the effectiveness of employing our MIM strategy for LLMs}, particularly in addressing the challenges of forgetting prevention and overfitting reduction. Does using LLM heads according to our strategy eliminate the prejudice about the unsuitability of LLMs in classification-based problems, specifically FCRE?
\end{itemize}



\paragraph{How to adapt BERT-based FCRE methods to LLMs?} Because current FCRE methods are used for BERT-based backbones, which are \textit{"encoder-only"} language models. It is essential to modify their original design to adapt to \textit{"decoder-only"} LLMs like LLAMA2-7B \citep{DBLP:journals/corr/abs-2307-09288}, Mistral-7B \citep{jiang2023mistral}, which operate in the auto-regressive mechanism \citep{DBLP:journals/corr/abs-1711-09534, DBLP:journals/corr/abs-1906-08237, touvron2023llama}. See illustration in Figure \ref{fig:llm_trans}, Appendix. In particular:

\begin{itemize}
    \item (i) The prompted inputs will be in the form of: \textit{"[Original sentence]. The relation between [Entity 1] and [Entity 2] \underline{is} [Answer]"};

    \item (ii) The embedding used for the main classifier (i.e., $g_\phi (\cdot)$) is now the embedding of the word \textit{\underline{"is"}} in the corresponding input, instead of "[MASK] embedding" in Figure \ref{fig:framework}. 
\end{itemize} 


\section{Experimental Results}

In this part, we first present the experiment setup in Section \ref{exp_setup}, followed by the results that demonstrate the effectiveness of our proposed method (Section \ref{exp_method}) when using BERT-based backbones. We then discuss the investigation results of using pre-trained LLMs for FCRE tasks in Section \ref{exp_LLM}.

\subsection{Experiment Setup}
\label{exp_setup}




In our experiments, {we use three current state-of-the-art methods as baselines, including:} SCKD \cite{DBLP:conf/acl/WangWH23}, ConPL \cite{DBLP:conf/acl/ChenWS23}, and CPL \citep{DBLP:conf/coling/MaHL024}. Besides, the models are evaluated using pre-trained models consisting of BERT \citep{DBLP:journals/corr/abs-1810-04805}, LLAMA2-7B \citep{DBLP:journals/corr/abs-2307-09288}, and Mistral-7B \citep{jiang2023mistral}, on two benchmark datasets: FewRel \cite{han-etal-2018-fewrel} and TACRED \cite{zhang-etal-2017-position}. 
We note that we have reproduced the results of ConPL \cite{DBLP:conf/acl/ChenWS23} under the same setting as SCKD and CPL. The reason is that the evaluation strategy in this paper is impractical for continual learning scenarios. Please refer to Appendix \ref{appx_implementation} for more details.

\subsection{Evaluation}
\label{exp_method}

\textbf{a. Using LM heads significantly improves the model's accuracy.} Table \ref{table:main1} reports the results of baselines and our proposed method \textbf{\textit{(+MI)}}, which exploits pre-trained LM heads beside the primary classifiers. In general, our method consistently helps improve the performance of existing methods in all cases. On both datasets, our strategy improved the final accuracy by around 2\% when integrated with CPL and ConPL and around 1\% when combined with SCKD. Moreover, considering accuracy after learning immediate tasks, ConPL+MI, when using our proposed strategy, can exceed the original version by about 15\% on TACRED.



\textbf{b. Exploiting the LM head effectively helps reduce forgetting and overfitting.} Figure \ref{fig:forget} and Table \ref{table:forgetting} show the accuracy drop after completing 8 tasks in various cases. The results indicate that our method significantly helps reduce forgetting for the baselines by approximately 1 to 3\%. 
Moreover, Figure \ref{fig:overfit} shows generalization gaps (i.e., $\delta = \text{test loss} - \text{train loss}$) after training each task of different models. The results show that our MIM strategy helps the models minimize these gaps significantly, thereby increasing their generalization.


\begin{table}[!ht]
    \centering
    
    \resizebox{.85\columnwidth}{!}{
    \begin{tabular}{lcgcg}
    \hline
    & \multicolumn{2}{c} {FewRel} & \multicolumn{2}{c} {TACRED} \\
    \cline{2-3} \cline{4-5} 
        & Original & + MI & Original & + MI \\ \hline
        SCKD & 36.31 & \textbf{33.92} & 31.77 & \textbf{30.80} \\ 
        ConPL & 37.80 & \textbf{35.12} & 32.72 & \textbf{30.52} \\ 
        CPL & 28.88 & \textbf{26.19} & 30.37 & \textbf{28.42} \\ \hline
    \end{tabular}}
    \caption{Accuracy drop (\%) after learning eight tasks of methods on the FewRel and TACRED in \textit{5-sho}t settings.}
    \label{table:forgetting}
\end{table}

\begin{table*}[!ht]
    \centering
    
    \begin{tabular}{lllllllll}
    \hline
        Method & $\mathcal{T}^1$ & $\mathcal{T}^2$ & $\mathcal{T}^3$ & $\mathcal{T}^4$ & $\mathcal{T}^5$ & $\mathcal{T}^6$ & $\mathcal{T}^7$ & $\mathcal{T}^8$ \\ \hline \hline
        \rowcolor{lightgray} ConPL + MI & 88.10 & \textbf{83.03} & 73.19 & 65.21 & 59.77 & \textbf{60.99} & \textbf{58.88} & \textbf{52.98} \\ 
        w.o MI (ConPL) & \textbf{88.77} & 69.64 & 57.50 & 52.15 & 58.19 & 55.01 & 52.88 & 50.97 \\
        w.o $L_{cc} \dagger$ & 88.15 & 83.01 & 73.11 & 65.16 & 58.70 & {60.06} & {58.61} & 52.69 \\ 
        w.o $L_{dc} \dagger$ & 88.10 & 82.67 & 73.10 & 65.06 & 58.70 & 60.36 & 58.71 & 52.84 \\ 
        w.o $L_{fc} \dagger$ & 88.06 & 81.15 & 72.04 & 63.15 & 56.26 & 59.30 & 57.69 & 50.10 \\ 
        freeze LM head & 88.22 & 80.27 & \textbf{77.15} & \textbf{67.72} & \textbf{59.62} & 57.75 & 54.73 & 52.10 \\ \hline \hline
        
        \rowcolor{lightgray} SCKD + MI & 87.55 & \textbf{79.39} & 70.70 & \textbf{66.78} & \textbf{61.94} & \textbf{59.81} & 55.10 & \textbf{53.63} \\
        w.o MI (SCKD) & \textbf{88.42} & {79.35} & 70.61 & {66.68} & 60.47 & 58.05 & 54.41 & 52.11 \\ 
        w.o $L_{dst} \dagger$ & 87.61 & 77.15 & 67.21 & 62.21 & 57.11 & 54.98 & 50.53 & 50.38 \\ 
        w.o $aug \dagger$ & 87.66 & 78.06 & 69.29 & 66.16 & {61.06} & 59.71 & 55.05 & 53.38 \\ 
        w.o $L_{dst}$ and $aug \dagger$ & 87.37 & 76.81 & 65.88 & 62.03 & 56.81 & 52.87 & 49.41 & 46.09 \\ 
        freeze LM head & 87.61 & 78.41 & \textbf{70.62} & 65.98 & 61.33 & 58.90 & \textbf{55.19} & 51.99 \\ \hline \hline
        
        \rowcolor{lightgray} CPL + MI & 85.67 & \textbf{82.54} & \textbf{75.12} & \textbf{70.65} & \textbf{66.79} & \textbf{65.17} & \textbf{61.25} & \textbf{59.48} \\
        freeze LM head & \textbf{86.17} & 80.52 & 73.84 & 69.03 & 64.33 & 62.36 & 60.19 & 57.99 \\ \hline
    \end{tabular}
    \caption{Ablation study on TACRED in the \textit{5-way-5-
shot} setting. $ \dagger$The components in the ablation study of the existing methods are described in Appendix \ref{sec:appendix_baseline}.}
    \label{abl}
    \vspace{-1mm}
\end{table*}

\textbf{c. The LM head supports representation learning.} Figure \ref{fig:overlap_main} presents representations in the latent space of CPL model before and after exploiting our MIM strategy (CPL+MI) on data of Task 1, after learning 8 tasks. It can be seen that the test features belonging to different categories of CPL+MI are better separated and therefore achieve better results.  
In addition, we provide a t-SNE visualization about features of the first task in the latent space on the LM head after learning the final tasks (Figure \ref{fig:tnse}), confirming the benefits when taking advantage of this component to enhance the performance of models. 
\begin{figure}[ht]
  \includegraphics[width=\columnwidth]{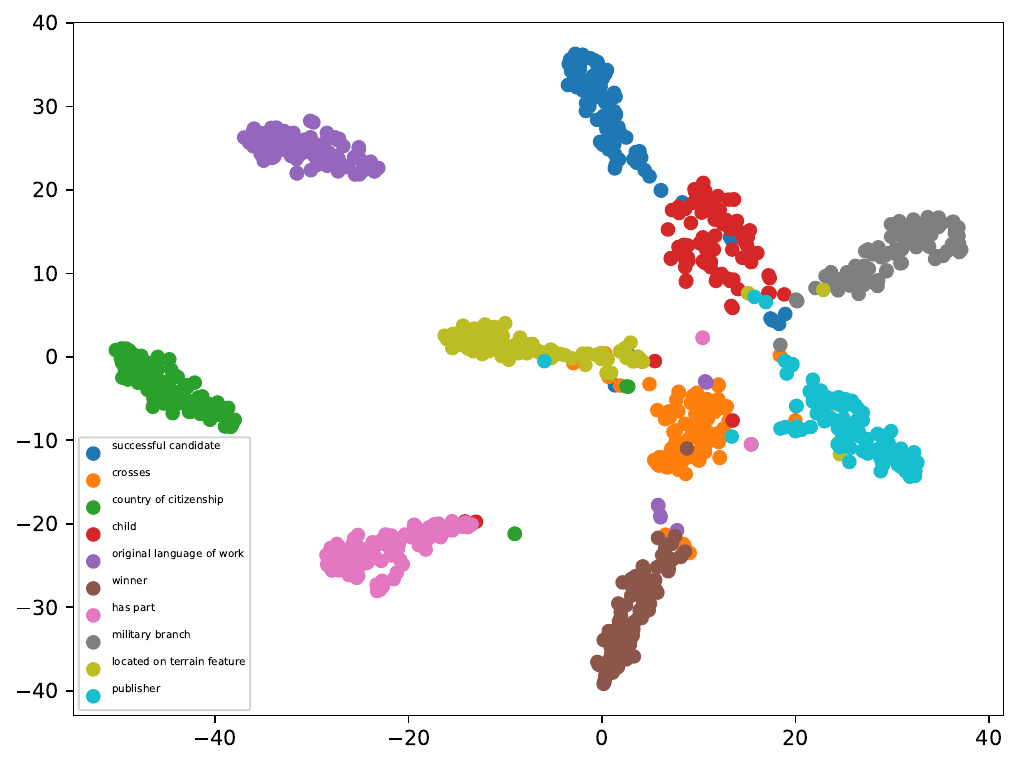}
  \caption{\textit{t}-SNE visualization of the representation of 10 relations from the first task of CPL+MI on the LM head after the last task (FewRel \textit{10-way 5-shot}).}
  \label{fig:tnse}

\end{figure}

\textbf{d. Ablation study} 

For further analyzing the effectiveness of our proposed method, we make an ablation study and present the experimental results in Table \ref{abl}. Regarding ConPL, it becomes evident that our MIM (e.i., +MI) plays a pivotal role compared to the loss components proposed in the original paper. Specifically, the elimination of each loss component among $L_{cc}$, $L_{dc}$ and $L_{fc}$ leads to only a marginal decline in performance. However, removing MI results in a notable decrease in accuracy across tasks, except for tasks 1 and 5. In the case of the SCKD, we note a substantial impact when excluding the distillation element (i.e., $L_{dst}$). This underscores the pivotal role of this component in mitigating forgetting while our proposed MI mechanism continues to enhance the performance of the overall model. 

\begin{table}[!ht]
    \centering

    \resizebox{1.0\columnwidth}{!}{
    \begin{tabular}{lcgcg}
    \hline
    & \multicolumn{2}{c} {FewRel} & \multicolumn{2}{c} {TACRED} \\
    \cline{2-3} \cline{4-5} 
        & Original & + MI & Original & + MI \\ \hline
        SCKD & 62.98 & \textbf{63.45} & 52.11 & \textbf{53.63} \\ 
        Llama2-7B-SCKD & 65.14 & \textbf{66.58} & 54.26 & \textbf{55.17}\\ \hline
        ConPL & 62.46 & \textbf{64.50} & 50.97 & \textbf{52.98}\\ 
        Llama2-7B-ConPL & 63.97 & \textbf{65.18} & 54.72 & \textbf{56.07} \\ \hline
        CPL & 64.50 & \textbf{66.27} & 57.39 & \textbf{59.48}\\ 
        Llama2-7B-CPL & 69.87 & \textbf{72.08} & 58.03 & \textbf{62.04} \\
        Mistral-7B-CPL & 71.89 & \textbf{75.02} & 64.11 & \textbf{65.48} \\ \hline
    \end{tabular}}
    \caption{Final Accuracy (\%) of methods after training the final task in the 5-shot settings.}
    \vspace{-3mm}
    \label{table:summary}
\end{table}

Moreover, we also explore a scenario in which the LM head is frozen to retain the knowledge from the pretraining phase fully. We notice inconsistent changes during the task learning process, with certain tasks demonstrating performance improvements while others exhibit declines. We hypothesize that in specific cases, the LM's pretraining-derived general knowledge can facilitate recognizing specific relations. Consequently, fine-tuning the model on domain-restricted data might compromise this capability. Conversely, for other relations, the general knowledge of the pretraining stage may not hold significant value.

\subsection{Using LLM for FCRE}
\label{exp_LLM}

\paragraph{\textbf{RQ1: How the performance would LLMs yield in FCRE tasks?}}

Table \ref{table:summary} depicts the increase in final accuracy after learning 8 FCRE tasks when the BERT-based backbone is replaced by the LLM backbone. Specifically, improvements can be as much as 3.75\% in the case of LLAMA-2-7B, and 8.75\% for Mistral-7B across both datasets. In addition, Table \ref{table:tacred_main} shows the full results of FCRE models on both datasets. Mostly, during the training of eight tasks, the LLMs tend to provide higher accuracy than the BERT-based models. For some immediate tasks, LLAMA2-7B can achieve up to 16\% higher accuracy than BERT-based models in TACRED, although their accuracy can be slightly lower in other cases. Besides, the differences in performance after training the first task and the last task (Accuracy drop - column $\Delta \downarrow$) in LLMs are smaller than in BERT-based models, from 2 to 5\% in the case of LLAMA2-7B and as much as 8\% for Mistral-CPL. These experimental results confirm the general superiority of LLM in solving FCRE compared to the class of conventional BERT-based models. 

\begin{table*}[!ht]
    \centering

    \resizebox{1.0\textwidth}{!}{
    \begin{tabular}{lllllllll|c}
        \textbf{FewRel} \textit{(10-way--5-shot)} \\
        \hline
        Method & $\mathcal{T}^1$ & $\mathcal{T}^2$ & $\mathcal{T}^3$ & $\mathcal{T}^4$ & $\mathcal{T}^5$ & $\mathcal{T}^6$ & $\mathcal{T}^7$ & $\mathcal{T}^8$ & $\Delta \downarrow$ \\ 
        \hline \hline
        SCKD & 94.75 & 82.83 & 76.21 & 72.19 & 70.61 & 67.15 & 64.86 & 62.98 & 31.77 \\ 
        
        \rowcolor{lightgray} SCKD + MI   & 94.75 & 83.88 & \textbf{76.71} & 72.34 & 70.78 & 67.36 & 65.08 & 63.45 & 31.30 \\ 
        Llama2-7B-SCKD & \textbf{95.63} & 82.76 & 76.04 & 74.91 & 70.10 & 66.52 & 64.89 & 65.14 & 30.49 \\ 
        
        \rowcolor{lightgray} Llama2-7B-SCKD + MI   & {95.22} & \textbf{85.01} & {76.63} & \textbf{76.50} & \textbf{72.19} & \textbf{67.47} & \textbf{67.03} & \textbf{66.58} & \textbf{28.64}\\ \hline \hline
        
        ConPL$^{**}$ & \textbf{95.18} & 79.63 & 74.54 & 71.27 & 68.35 & 63.86 & 64.74 & 62.46 &32.72 \\ 
        
        \rowcolor{lightgray} ConPL + MI   & 95.02 & 81.42 & 77.23 & 74.21 & 69.64 & 67.74 & \textbf{66.44} & 64.50 & 30.52\\ 
        
        Llama2-7B-ConPL & 94.72 & 82.43 & 75.07 & 73.95 & 72.67 & 65.80 & 63.41 & 63.79& 30.93 \\ 
        
        \rowcolor{lightgray} Llama2-7B-ConPL + MI   & 94.50 & \textbf{83.75} & \textbf{77.61} & \textbf{74.78} & \textbf{72.83} & \textbf{68.01} & {63.98} & \textbf{65.18} & \textbf{29.32}\\ \hline \hline
        
        CPL & 94.87 & 85.14 & 78.80 & 75.10 & 72.57 & 69.57 & 66.85 & 64.50 & 30.37\\ 
        \rowcolor{lightgray} CPL + MI   & 94.69 & 85.58 & 80.12 & 75.71 & 73.90 & 70.72 & 68.42 & 66.27 & 28.42\\ 
        
        Llama2-7B-CPL & 95.73 & 85.87 & 80.57 & 78.60 & 77.30 & 73.95 & 71.35 & 69.87 & 25.86\\
        \rowcolor{lightgray} Llama2-7B-CPL + MI   & 95.63 & 87.14 & 83.25 & 80.59 & \textbf{79.20} & 76.41 & 74.62 & 72.08 & 23.55 \\ 
        Mistral-7B-CPL & \textbf{96.57} & 86.80 & 83.31 & 79.45 & 77.17 & 74.24 & 73.59 & 71.89 & 24.68\\
        \rowcolor{lightgray} Mistral-7B-CPL + MI   & 96.55 & \textbf{90.77} & \textbf{84.81} & \textbf{83.08} & {78.92} & \textbf{77.27} & \textbf{77.05} & \textbf{75.02} & \textbf{21.53}\\ \hline \\
        \textbf{TACRED} \textit{(5-way-5-shot)} \\
    \hline
        Method & $\mathcal{T}^1$ & $\mathcal{T}^2$ & $\mathcal{T}^3$ & $\mathcal{T}^4$ & $\mathcal{T}^5$ & $\mathcal{T}^6$ & $\mathcal{T}^7$ & $\mathcal{T}^8$ & $\Delta \downarrow$ \\ \hline
        \hline
        SCKD & 88.42 & 79.35 & 70.61 & 66.78 & 60.47 & 58.05 & 54.41 & 52.11 & 36.31\\ 
        
        \rowcolor{lightgray} SCKD + MI   & 87.55 & 78.39 & 69.70 & \textbf{66.88} & 61.94 & 59.81 & 55.10 & 53.63 & 33.92\\ 
        
        Llama2-7B-SCKD & \textbf{88.67} & 84.48 & 72.53 & 63.10 & 62.01 & 59.38 & 57.18 & 54.26 & 34.41\\ 
        
        \rowcolor{lightgray} Llama2-7B-SCKD + MI & 88.35 & \textbf{84.90} & \textbf{74.32} & 63.48 & \textbf{63.37} & \textbf{60.20} & \textbf{59.64} & \textbf{55.17} & \textbf{33.18}\\ \hline \hline
        
        ConPL$^{**}$ & \textbf{88.77} & 69.64 & 57.50 & 52.15 & 58.19 & 55.01 & 52.88 & 50.97 & 37.80\\ 
        
        \rowcolor{lightgray} ConPL + MI   & 88.10 & 83.03 & 73.19 & 65.21 & 58.77 & \textbf{60.99} & \textbf{58.88} & 52.98 & 35.12 \\ 
        Llama2-7B-ConPL & 87.26 & 81.72 & 73.04 & 65.67 & 60.96 & 58.47 & 56.49 & 54.72 & 32.54\\
        
        \rowcolor{lightgray} Llama2-7B-ConPL + MI & 86.88 & \textbf{83.11} & \textbf{73.83} & \textbf{67.58} & \textbf{61.87} & {60.31} & 56.83 & \textbf{56.07} & \textbf{30.81}\\ \hline \hline
        CPL & 86.27 & 81.55 & 73.52 & 68.96 & 63.96 & 62.66 & 59.96 & 57.39 & 28.88\\ 
        
        \rowcolor{lightgray} CPL + MI   & 85.67 & \textbf{82.54} & 75.12 & 70.65 & 66.79 & 65.17 & 61.25 & 59.48 & 26.19 \\

        Llama2-7B-CPL & \textbf{86.76} & 75.94 & 70.65 & 68.64 & 67.44 & 65.12 & 60.27 & 58.03 & 30.23\\
        \rowcolor{lightgray} Llama2-7B-CPL + MI   & 85.55 & 77.91 & 76.49 & 74.99 & 69.15 & 68.19 & 64.19 & 62.04 & 23.51\\ 
        Mistral-7B-CPL & 86.67 & 80.98 & 77.16 & 73.24 & 70.05 & 67.70 & 67.04 & 64.11 & 22.56\\
        \rowcolor{lightgray} Mistral-7B-CPL + MI   & 86.32 & {81.00} & \textbf{77.71} & \textbf{75.48} & \textbf{71.92} & \textbf{71.02} & \textbf{67.69} & \textbf{65.48} &  \textbf{20.84} \\ \hline \\
    \end{tabular}}
    \vspace{-4mm}
    \caption{Accuracy (\%) of methods using different LMs after training for each task. We \colorbox{lightgray}{highlight} the rows corresponding to our proposed method. The best result in each group is in \textbf{bold}. **Results of ConPL are reproduced. Columns $\Delta \downarrow$ present Accuracy drop after learning 8 tasks.}
    \label{table:tacred_main}
\end{table*}

On the other hand, pre-trained LLMs are known to be knowledge-rich models with high generalization capabilities. However, for the first task, LLMs achieve accuracies of around 96\% on FewRel and around 86\% on TACRED, having no clear advantage over BERT-based models. Besides, the results in Table \ref{table:tacred_main} clearly demonstrate the degradation of prior knowledge when applying pre-trained LLM in FCRE.  In particular, the model's accuracy can drop by 30 - 32\% for LLAMA2-7B and by 20 - 25\% for Mistral-7B, after training 8 tasks. 

Thanks to thorough training on large datasets, LLMs with billions of parameters contain a wealth of knowledge and have great potential in downstream tasks. However, in some cases, with the current operating mechanism of an autoregressive decoder, employing such a model with billions of parameters, as opposed to one with hundreds of millions (BERT), proves exceedingly expensive for only marginal improvements in accuracy. Even on TACRED, the final accuracy of LLAMA2-7B-CPL is lower than that of CPL+MI, indicating that our method with the BERT-based model can effectively replace the LLM in this case. These findings necessitate the development of more effective methodologies to ensure the effectiveness of LLMs within this challenging setting


\paragraph{\textbf{RQ2: The effectiveness of exploiting our MIM strategy for LLMs in FCRE tasks}}

Figure \ref{fig:forget} and Table \ref{table:tacred_main} clearly show that our strategy significantly mitigates accuracy drop in LLMs, which could reach up to 6\% on TACRED and 4\% on FewRel, and better than on BERT-based models. Besides, Figure \ref{fig:overfit} consistently illustrates the effectiveness of our method in reducing overfitting. 
It can be said that with our proposed strategy, LLM heads are no longer an obstacle when applying pre-trained LLMs to classification tasks. On the contrary, using LLMs demonstrates the clearest and most significant improvement in mitigating catastrophic forgetting and reducing overfitting.



\section{Conclusion}


In this work, we introduce a novel method that utilizes pre-trained language model heads to maintain the generalization of LMs in FCRE problems. By making use of this often ignored component through a mutual information strategy, our approach also significantly improves the comprehensiveness of the representation on the main classifier. Additionally, we present comprehensive experimental results that demonstrate the impact of using LLMs for FCRE and provide valuable insights to the community.

\section*{Limitations}

\begin{itemize}
    \item First, our proposed method and current investigations in this paper apply only to high-level RE tasks, where all entities are assumed to be given. Therefore, to achieve more practical results, it is motivating to consider end-to-end RE problems, covering entity recognition to relation extraction between entities in the future.
    


    \item Another potential limitation could arise from the fact that pre-trained LMs used in our work might inherit biases from their pre-training data. These biases can manifest in various forms, such as gender, racial, or cultural biases, and could be exacerbated in scenarios with limited labeled data, as in FCRE tasks. Our method endeavors to transfer the knowledge within the LMs to the classification head by leveraging Mutual Information (MI), which could inadvertently perpetuate biased representations. Such biased representations may have adverse consequences, potentially resulting in misidentifying relations associated with biased information. This raises an open question for the research community to investigate further, exploring the impact of bias on FCRE tasks when utilizing LLMs.
\end{itemize}

\section*{Acknowledgements}

This research has been supported by the Army Research Office (ARO) grant W911NF-21-1-0112, the NSF grant CNS-1747798 to the IUCRC Center for Big Learning, and the NSF grant \# 2239570. This research is also supported in part by the Office of the Director of National Intelligence (ODNI), Intelligence Advanced Research Projects Activity (IARPA), via the HIATUS Program contract 2022-22072200003. The views and conclusions contained herein are those of the authors and should not be interpreted as necessarily representing the official policies, either expressed or implied, of ODNI, IARPA, or the U.S. Government.

\bibliography{main}


\clearpage
\appendix
\label{sec:appendix}

\begin{table*}[!ht]
    \centering

    \resizebox{1.0\textwidth}{!}{
    \begin{tabular}{lllllllll|c}
        \textbf{FewRel} \textit{(10-way--5-shot)} \\
        \hline
        Method & $\mathcal{T}^1$ & $\mathcal{T}^2$ & $\mathcal{T}^3$ & $\mathcal{T}^4$ & $\mathcal{T}^5$ & $\mathcal{T}^6$ & $\mathcal{T}^7$ & $\mathcal{T}^8$ & $\Delta \downarrow$ \\ 
        \hline \hline
        SCKD & 94.75 & 82.83 & 76.21 & 72.19 & 70.61 & 67.15 & 64.86 & 62.98 & 31.77 \\ 
        
        \rowcolor{lightgray} SCKD + MI   & $94.75_{\pm 0.37}$ & $83.88_{\pm 0.67}$ & $76.71_{\pm 2.48}$ & $72.34_{\pm 1.43}$ & $70.78_{\pm 0.82}$ & $67.36_{\pm 0.73}$ & $65.08_{\pm 2.43}$ & $63.45_{\pm 2.44}$ & 31.30 \\ 

        Llama2-7B-SCKD & \textbf{$95.63_{\pm 0.56}$} & $82.76_{\pm 2.26}$ & $76.04_{\pm 4.22}$ & $74.91_{\pm 
        .77}$ & $70.10_{\pm 3.63}$ & $66.52_{\pm 2.9}$ & $64.89_{\ 2.85}$ & $65.14_{\pm 1.52}$ & 30.49 \\ 

        \rowcolor{lightgray} Llama2-7B-SCKD + MI   & {$95.22_{\pm 0.53}$} & \textbf{$85.01_{\pm 2.4}$} & {$76.63_{\pm 1.19}$} & \textbf{$76.50_{\pm 1.28}$} & \textbf{$72.19_{\pm 1.4}$} & \textbf{$67.47_{\pm 1.87}$} & \textbf{$67.03_{\pm 2.97}$} & \textbf{$66.58_{\pm 2.11}$} & \textbf{28.64}\\ \hline \hline

        ConPL$^{**}$ & $95.18_{\pm 0.73}$ & $79.63_{\pm 1.27}$ & $74.54_{\pm 1.13}$ & $71.27_{\pm 0.85}$ & $68.35_{\pm 0.86}$ & $63.86_{\pm 2.03}$ & $64.74_{\pm 1.39}$ & $62.46_{\pm 1.54}$ &32.72 \\ 
        
        \rowcolor{lightgray} ConPL + MI & $95.02_{\pm 0.4}$ & $81.42_{\pm 1.93}$ & $77.23_{\pm 1.01}$ & $74.21_{\pm 1.5}$ & $69.64_{\pm 1.19}$ & $67.74_{\pm 1.52}$ & $66.44_{\pm 1.91}$ & $64.50_{\pm 1.15}$ & 30.52\\ 
        
        Llama2-7B-ConPL & $94.72_{\pm 1.15}$ & $82.43_{\pm 1.69}$ & $75.07_{\pm 1.62}$ & $73.95_{\pm 2.75}$ & $72.67_{\pm 1.51}$ & $65.80_{\pm 1.46}$ & $63.41_{\pm 2.15}$ & $63.79_{\pm 2.76}$ & 30.93 \\ 
        
        \rowcolor{lightgray} Llama2-7B-ConPL + MI   & $94.50_{\pm 0.57}$ & \textbf{$83.75_{\pm 1.05}$} & \textbf{$77.61_{\pm 1.27}$} & \textbf{$74.78_{\pm 3.19}$} & \textbf{$72.83_{\pm 2.74}$} & \textbf{$68.01_{\pm 2.23}$} & {$63.98_{\pm }3.1$} & \textbf{$65.18_{\pm 1.99}$} & \textbf{29.32}\\ \hline \hline
        
        CPL & 94.87 & 85.14 & 78.80 & 75.10 & 72.57 & 69.57 & 66.85 & 64.50 & 30.37\\ 
        \rowcolor{lightgray} CPL + MI  & $94.69_{\pm 0.7}$ & $85.58_{\pm 1.88}$ & $80.12_{\pm 2.45}$ & $75.71_{\pm 2.28}$ & $73.90_{\pm 1.8}$ & $70.72_{\pm 0.91}$ & $68.42_{\pm 1.77}$ & $66.27_{\pm 1.58}$ & 28.42\\ 
        
        Llama2-7B-CPL & $95.73_{\pm 0.92}$ & $85.87_{\pm 1.46}$ & $80.57_{\pm 1.74}$ & $78.60_{\pm 3.31}$ & $77.30_{\pm 2.41}$ & $73.95_{\pm 1.54}$ & $71.35_{\pm 3.75}$ & $69.87_{\pm 2.32}$ & 25.86\\
        
        \rowcolor{lightgray} Llama2-7B-CPL + MI   & $95.63_{\pm 1.08}$ & $87.14_{\pm 1.94}$ & $83.25_{\pm 2.14}$ & $80.59_{\pm 2.37}$ & \textbf{$79.20_{\pm 1.36}$} & $76.41_{\pm 2.13}$ & $74.62_{\pm }1.73$ & $72.08_{\pm 3.18}$ & 23.55 \\ 

        Mistral-7B-CPL & \textbf{$96.57_{\pm 0.40}$} & $86.80_{\pm 2.53}$ & $83.31_{\pm 1.94}$ & $79.45_{\pm 2.53}$ & $77.17_{\pm 2.2}$ & $74.24_{\pm 1.96}$ & $73.59_{\pm 2.00}$ & $71.89_{\pm 1.97}$ & 24.68\\

        \rowcolor{lightgray} Mistral-7B-CPL + MI   & $96.55_{\pm 0.43}$ & \textbf{$90.77_{\pm 2.11}$} & \textbf{$84.81_{\pm 1.09}$} & \textbf{$83.08_{\pm 1.5}$} & {$78.92_{\pm 1.35}$} & \textbf{$77.27_{\pm 2.06}$} & \textbf{$77.05_{\pm 2.3}$} & \textbf{$75.02_{\pm 1.67}$} & \textbf{21.53}\\ \hline \\
        \textbf{TACRED} \textit{(5-way-5-shot)} \\
    \hline
        Method & $\mathcal{T}^1$ & $\mathcal{T}^2$ & $\mathcal{T}^3$ & $\mathcal{T}^4$ & $\mathcal{T}^5$ & $\mathcal{T}^6$ & $\mathcal{T}^7$ & $\mathcal{T}^8$ & $\Delta \downarrow$ \\ \hline
        \hline
        SCKD & 88.42 & 79.35 & 70.61 & 66.78 & 60.47 & 58.05 & 54.41 & 52.11 & 36.31\\ 
        
        \rowcolor{lightgray} SCKD + MI   & $87.55_{\pm 0.48}$ & $78.39_{\pm 2.18}$ & $69.70_{\pm 1.75}$ & $66.88_{\pm 1.56}$ & $61.94_{\pm 2.87}$ & $59.81_{\pm 1.56}$ & $55.10_{\pm 3.63}$ & $53.63_{\pm 2.31}$ & 33.92\\ 
        
        Llama2-7B-SCKD & \textbf{$88.67_{\pm 0.56}$} & $84.48_{\pm 2.26}$ & $72.53_{\pm 4.22}$ & $63.10_{\pm 4.77}$ & $62.01_{\pm 3.63}$ & $59.38_{\pm 2.90}$ & $57.18_{\pm 2.85}$ & $54.26_{\pm 1.52}$ & 34.41\\ 
        
        \rowcolor{lightgray} Llama2-7B-SCKD + MI & $88.35_{\pm 1.11}$ & \textbf{$84.90_{\pm 2.59}$} & \textbf{$74.32_{\pm 3.73}$} & $63.48_{\pm 2.03}$ & \textbf{$63.37_{\pm 2.44}$} & \textbf{$60.20_{\pm 3.54}$} & \textbf{$59.64_{\pm 3.19}$} & \textbf{$55.17_{\pm 2.68}$} & \textbf{33.18}\\ \hline \hline

        ConPL$^{**}$ & $88.77_{\pm 0.84}$ & $69.64_{\pm 1.93}$ & $57.50_{\pm 2.48}$ & $52.15_{\pm 1.59}$ & $58.19_{\pm 2.31}$ & $55.01_{\pm 3.12}$ & $52.88_{\pm 3.66}$ & $50.97_{\pm 3.41}$ & 37.80\\ 
        
        \rowcolor{lightgray} ConPL + MI   & $88.10_{\pm 0.68}$ & $83.03_{\pm 3.38}$ & $73.19_{\pm 1.57}$ & $65.21_{\pm 3.04}$ & $58.77_{\pm 3.45}$ & $60.99_{\pm 1.61}$ & $58.88_{\pm 2.52}$ & $52.98_{\pm 1.68}$ & 35.12 \\ 

        Llama2-7B-ConPL & $87.26_{\pm 1.22}$ & $81.72_{\pm 2.54}$ & $73.04_{\pm 2.92}$ & $65.67_{\pm 2.07}$ & $60.96_{\pm 4.39}$ & $58.47_{\pm 3.32}$ & $56.49_{\pm 3.2}$ & $54.72_{\pm 2.24}$ & 32.54\\
        
        \rowcolor{lightgray} Llama2-7B-ConPL + MI & $86.88_{\pm 1.03}$ & \textbf{$83.11_{\pm 3.46}$} & \textbf{$73.83_{\pm 2.88}$} & \textbf{$67.58_{\pm 2.04}$} & \textbf{$61.87_{\pm 4.16}$} & {$60.31_{\pm 4.41}$} & $56.83_{\pm 2.57}$ & \textbf{$56.07_{\pm 3.45}$} & \textbf{30.81}\\ \hline \hline
        CPL & 86.27 & 81.55 & 73.52 & 68.96 & 63.96 & 62.66 & 59.96 & 57.39 & 28.88\\ 
        
        \rowcolor{lightgray} CPL + MI   & $85.67_{\pm 0.8}$ & $82.54_{\pm 2.98}$ & $75.12_{\pm 3.67}$ & $70.65_{\pm 2.75}$ & $66.79_{\pm 2.18}$ & $65.17_{\pm 2.48}$ & $61.25_{\pm 1.52}$ & $59.48_{\pm 3.53}$ & 26.19 \\ 
        
        Llama2-7B-CPL & \textbf{$86.76_{\pm 1.58}$} & $75.94_{\pm 4.76}$ & $70.65_{\pm 2.57}$ & $68.64_{\pm 3.03}$ & $67.44_{\pm 2.95}$ & $65.12_{\pm 3.85}$ & $60.27_{\pm 3.79}$ & $58.03_{\pm 1.98}$ & 30.23\\

        \rowcolor{lightgray} Llama2-7B-CPL + MI   & $85.55_{\pm 0.74}$ & $77.91_{\pm 2.8}$ & $76.49_{\pm 2.79}$ & $74.99_{\pm 2.69}$ & $69.15_{\pm 3.65}$ & $68.19_{\pm 2.29}$ & $64.19_{\pm 3.01}$ & $62.04_{\pm 1.1}$ & 23.51\\ 
        
        Mistral-7B-CPL & $86.67_{\pm 0.81}$ & $80.98_{\pm 5.42}$ & $77.16_{\pm 4.96}$ & $73.24_{\pm 3.63}$ & $70.05_{\pm 2.5}$ & $67.70_{\pm 3.95}$ & $67.04_{\pm 3.12}$ & $64.11_{\pm 3.68}$ & 22.56\\
        
        \rowcolor{lightgray} Mistral-7B-CPL + MI   & $86.32_{\pm 1.25}$ & {$81.00_{\pm 3.2}$} & \textbf{$77.71_{\pm 2.31}$} & \textbf{$75.48_{\pm 2.59}$} & \textbf{$71.92_{\pm 3.09}$} & \textbf{$71.02_{\pm 2.84}$} & \textbf{$67.69_{\pm 3.58}$} & \textbf{$65.48_{\pm 1.97}$} &  \textbf{20.84} \\ \hline \\
    \end{tabular}}
    \vspace{-4mm}
    \caption{Accuracy (\%) of methods using different LMs after training for each task. We \colorbox{lightgray}{highlight} the rows corresponding to our proposed method. The best result in each group is in \textbf{bold}. **Results of ConPL are reproduced. Columns $\Delta \downarrow$ present Accuracy drop after learning 8 tasks.}
    \label{table:tacred_main}
\end{table*}

\begin{figure*}
    \centering
    \includegraphics[width=\textwidth]{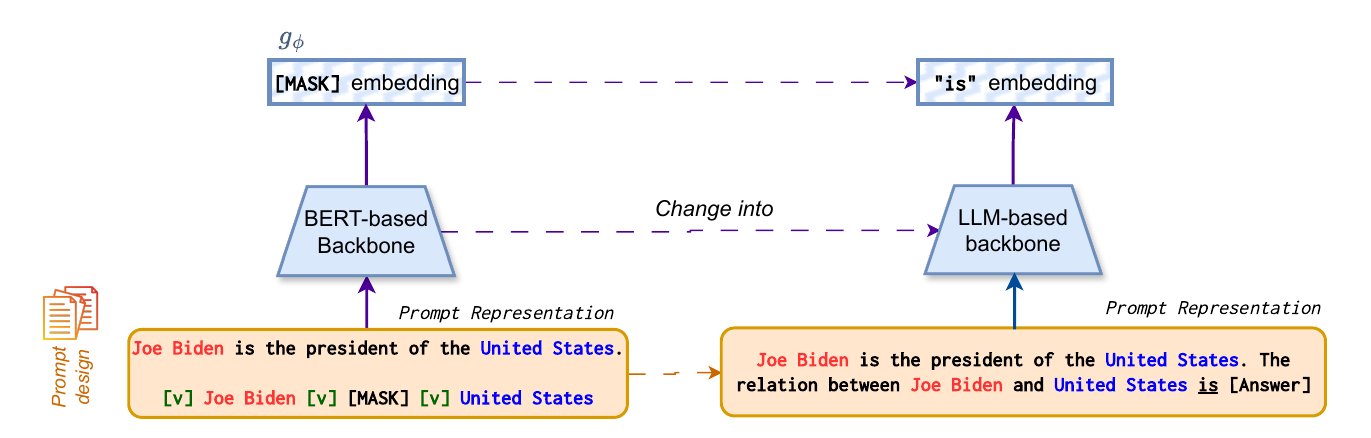}
    \caption{Adapting LLMs for FCRE problems}
    \label{fig:llm_trans}
\end{figure*}


\section{Implementation details}
\label{appx_implementation}

For each reported result, we conduct 6 independent runs with different random seeds and report the mean. Our code is available at \href{https://github.com/thanhnx12/CRE-via-MMI}{https://github.com/thanhnx12/CRE-via-MMI} \\

\textit{Note:} As discussed in \cite{DBLP:journals/corr/abs-2310-01208}, LLaMA-2-7B model gives better results compared with LLaMA-2-13B. Therefore, we opt to use LLaMA-2-7B to examine in our experiments.


\subsection{Datasets}
Our experiments utilize the following two benchmarks:
\begin{itemize}
    \item \textbf{FewRel} \cite{han-etal-2018-fewrel} includes 100 relations with 70,000 samples. Following \citet{DBLP:conf/acl/QinJ22}, we employ a setup with 80 relations, partitioned into 8 tasks, each comprising 10 relations \textit{(10-way)}. Task $\mathcal{T}^1$ includes 100 samples per relation, whereas the remaining tasks are characterized as few-shot tasks conducted under \textit{5-shot} settings.

    \item \textbf{TACRED} \cite{zhang-etal-2017-position} encompasses 42 relations with 106,264 samples extracted from Newswire and Web documents. Consistent with the approach outlined by \citet{DBLP:conf/acl/QinJ22}, we exclude instances labeled as "no\_relation" and allocate the remaining 41 relations across 8 tasks. Task $\mathcal{T}^1$ comprises 6 relations, each with 100 samples, while each subsequent tasks involve 5 relations \textit{(5-way)} in \textit{5-shot} setups.
    
\end{itemize}

\subsection{Baselines}
\label{sec:appendix_baseline}
In this work, we showcase our approach through thorough experiments using three recent SOTA methods in FCRE as the baselines, including:
\begin{itemize}
    \item \textbf{SCKD} \citep{DBLP:conf/acl/WangWH23}: adopts a systematic strategy for knowledge distillation, which aims to preserve old knowledge from previous tasks. Besides, this method employs contrastive learning techniques with pseudo samples to enhance the distinguishability between representations of different relations.

    In this paper, to conduct the ablation study in Table \ref{abl}, we denote $L_{dst}$ as the representative of all the losses serving the distillation and contrastive learning mentioned above and $aug$ as the augmentation technique on the memory buffer.

    \item \textbf{ConPL} \citep{DBLP:conf/acl/ChenWS23} proposes a method that consists of three fundamental modules: a prototype-based classification module, a memory-enhanced module, and a novel consistent learning module that enforces distribution consistency to prevent forgetting. Additionally, ConPL leverages prompt learning to improve representation learning and incorporate focal loss to alleviate confusion among closely related classes.

    This paper conducts the ablation study in Table \ref{abl}where the role of each component of ConPL's objective function is analyzed. In particular, $L_{cc}$ helps constrain the consistency between samples and corresponding prototypes of old tasks, $L_{dc}$ forces the consistency regarding the distribution of samples and prototypes, and $L_{fc}$ is a focal loss that alleviates the difficulty of choosing negative classes during inference. 

    \item \textbf{CPL} \cite{DBLP:conf/coling/MaHL024} CPL proposes a Contrastive Prompt Learning framework, which designs prompts to generalize across categories and uses margin-based contrastive learning to handle hard samples, thus reducing catastrophic forgetting and overfitting. Besides, the authors employ a memory augmentation strategy to generate diverse samples with ChatGPT, further mitigating overfitting in low-resource scenarios of FCRE.
\end{itemize}

\subsection{Evaluation Protocol}
\label{appx:eval}

\paragraph{Metric} We use final average accuracy to evaluate methods in our experiments. The average accuracy at task \( T_j \) is calculated as follows:

    \[ ACC_j = \frac{1}{j} \sum_{i=1}^{j} ACC_{j,i} \]

    where \( ACC_{j,i} \) is the accuracy on the test set of task \( T_i \) after training the model on task \( T_j \). 

\paragraph{Prediction mechanism} As mentioned in \ref{exp_setup}, our methods follow the evaluation strategy in the setting of SCKD and CPL. Specifically, during the testing phase, the learned model is required to evaluate all classes/ relations it has been trained on so far. 

Note that in the original code repository of ConPL (e.g., Lines 18-53 in this \href{https://github.com/XiudiChen/ConPL/blob/main/fewrel_5shot.py}{file}), this method follows a different evaluation process. In particular, after training on task $\mathcal{T}^k$, the model has been trained on a set of $\tilde{R^t}$ relations. However, for each relation $r$, ConPL defines a set of negative candidate classes $M_r$, so that predictions are made on the set $(\tilde{R^t} \cap M_r)$. This means that the model does not make predictions with all the classes it has learned so far but rather with a predefined subset specific to each relation. While enhancing the performance reported for ConPL, this targeted prediction approach does not align with the practical requirements of CL. In this challenging scenario, each model has to dynamically adapt and make predictions across the expanding set of relations without relying on some fixed set of classes. Therefore, despite its efficacy in controlled evaluations, the ConPL method is impractical for real-world continual learning applications.

\end{document}